\documentclass[10pt,twocolumn,letterpaper]{article}

\usepackage{iccv}
\usepackage{times}
\usepackage{epsfig}
\usepackage{graphicx}
\usepackage{amsmath}
\usepackage{amssymb}

\usepackage{overpic}
\usepackage{color}
\usepackage{epstopdf}
\usepackage{amsthm}
\usepackage{algorithm,algorithmic}
\usepackage{amsfonts}
\usepackage{multirow,array,bm}
\usepackage{helvet} 
\usepackage{courier}  

\usepackage[accsupp]{axessibility}

\usepackage[breaklinks=true,colorlinks,bookmarks=False]{hyperref}

\iccvfinalcopy 

\def\iccvPaperID{****} 

\ificcvfinal\fi

\begin{document}

\title{Variational Attention: Propagating Domain-Specific Knowledge for Multi-Domain Learning in Crowd Counting}

\author{Binghui Chen\thanks{Equal contribution}, Zhaoyi Yan$^{*1}$, Ke Li, Pengyu Li, Biao Wang, Wangmeng Zuo$^{1}$\thanks{Corresponding author}, Lei Zhang$^{2}$\\
$^{1}$ Harbin Institute of Technology, $^{2}$ The Hong Kong Polytechnic University\\
{\tt\small chenbinghui@bupt.cn, yanzhaoyi@outlook.com, like1990@bupt.edu.cn, lipengyu007@gmail.com}\\
{\tt\small wangbiao225@foxmail.com,wmzuo@hit.edu.cn,cslzhang@comp.polyu.edu.hk}
}

\maketitle
\ificcvfinal\thispagestyle{empty}\fi

\begin{abstract}
   In crowd counting, due to the problem of laborious labelling, it is perceived intractability of collecting a new large-scale dataset which has plentiful images with large diversity in density, scene, etc. Thus, for learning a general model, training with data from multiple different datasets might be a remedy and be of great value. In this paper, we resort to the multi-domain joint learning and propose a simple but effective Domain-specific Knowledge Propagating Network (DKPNet)\footnote{Code will be available at https://github.com/Zhaoyi-Yan/DKPNet} for unbiasedly learning the knowledge from multiple diverse data domains at the same time. It is mainly achieved by proposing the novel Variational Attention(VA) technique for explicitly modeling the attention distributions for different domains. And as an extension to VA, Intrinsic Variational Attention(InVA) is proposed to handle the problems of over-lapped domains and sub-domains. Extensive experiments have been conducted to validate the superiority of our DKPNet over several popular datasets, including ShanghaiTech A/B, UCF-QNRF and NWPU.
\end{abstract}

\vspace{-1em}
\section{Introduction}\label{sec_intro}

\begin{figure}[!t]
  \centering\vspace{-2em}
  \includegraphics[width=0.8\linewidth]{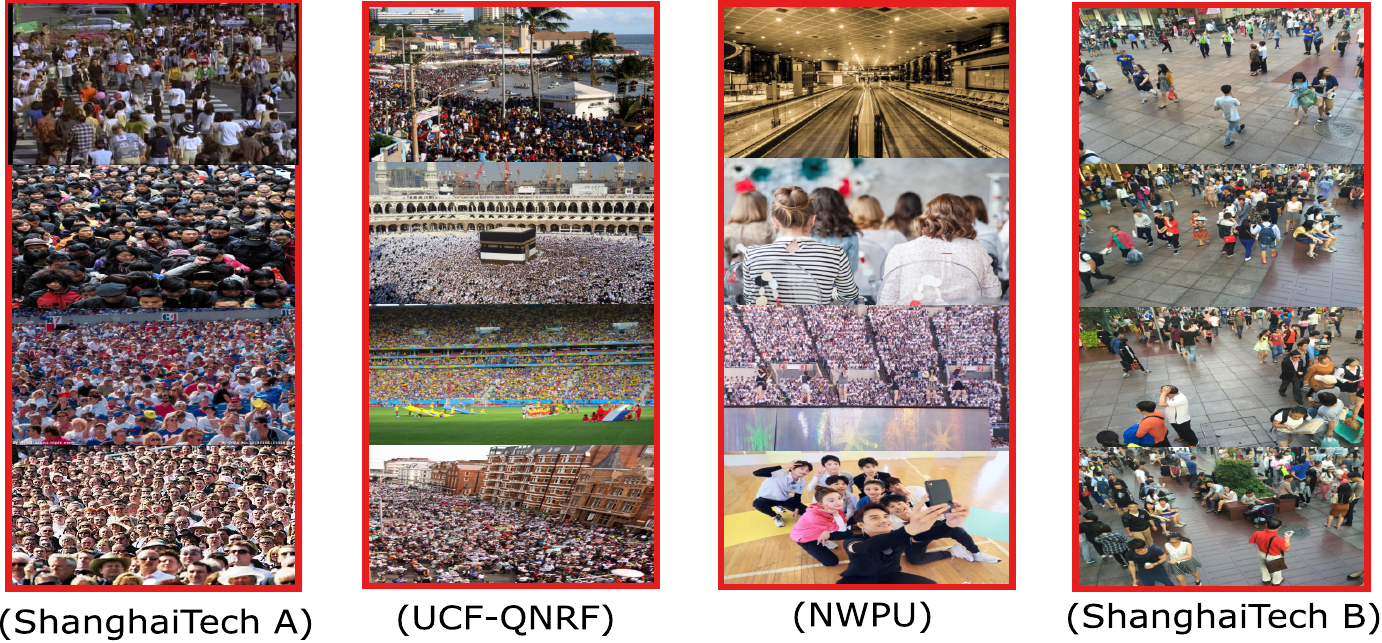}\\
  \caption{Data distribution comparison between ShanghaiTech \cite{zhang2016single}, UCF-QNRF \cite{idrees2018composition} and NWPU \cite{wang2020nwpu}. ShanghaiTech A is mainly composed of congested images, QNRF is of highly-congested samples and have more background scenarios, NWPU covers a much larger variety of data distributions due to density, perspective, background, \emph{etc}, while ShanghaiTech B prefers low density and ordinary street-based scenes.}\label{fig_intro}
  \vspace{-1.7em}
\end{figure}

Crowd counting is a challenging problem since it suffers from multiple actual issues behind data distributions, such as high variability in scales, density, occlusions, perspective distortions, background scenarios, \emph{etc}. A direct solution to mitigate these issues is to collect a large-scale dataset with abundant data variations like ImageNet\cite{deng2009imagenet}, so as to encourage the learned model to be more robust and general. However, collecting such a large-scale dataset with rich diversity for crowd-counting training is intractable due to the difficulty in human-labeling. Specifically, in crowd counting, due to the limitation of various conditions, images collected by a research group might only contain certain types of variations and are limited in numbers. For example, as shown in Fig.\ref{fig_intro}, one can observe that there are large variations in data distributions across different datasets. Images in ShanghaiTech A(SHA)\cite{zhang2016single} tend to show congested crowds, and those in UCF-QNRF(QNRF)\cite{idrees2018composition} are more likely to depict highly-congested crowds and have more background scenarios, and those in NWPU\cite{wang2020nwpu} have much more diversities in scales, density, background, \emph{etc}. In contrast, samples within ShanghaiTech B(SHB)\cite{zhang2016single} just prefer low density crowds and the ordinary street-based scenes. Considering the aforementioned facts, in order to learn a general and robust estimating model for correct density prediction, this paper resorts to \emph{multi-domain learning} which aims to solve the same or similar problem with multiple datasets across different domains\footnote{Commonly, a \emph{domain} often refers to a data set where samples follow the similar or same underlying data distribution\cite{xiao2016learning}.} simultaneously by utilizing all the data these domains provide. In other words, multi-domain learning gives chances of using relatively abundant data variations coming from different datasets for learning a general and robust density estimating model.

However, in crowd counting, an interesting phenomenon can be observed when jointly training with multiple different datasets. As shown in Tab.\ref{tab:JT}, if under the supervision of the 3-joint of SHA, SHB and QNRF, the deep model prefers to only improve the performances on SHA and QNRF, while sacrificing that on SHB ($5\%$ performance drop ). Actually, this kind of phenomenon (i.e. biased/partial learning) exists widely in computer vision fields. It is because that deep models have the partial/biased learning behavior\cite{chen2019energy}, i.e. \emph{deep models easily learn to focus on surface statistical regularities rather than more general abstract concepts}. In other words, deep models will selectively learn the dominant data knowledge from certain dominant domains\footnote{Since SHA and QNRF data are much more similar than SHB data, and when combining them three together, SHA and QNRF turn to be the dominant domains.} while ignoring other potential helpful information from the rest domains. To this end, developing an effective algorithm that can successfully utilize all the knowledge from different datasets remains important.
\begin{table}[!tbp]
  \centering\centering
  \caption{MAE Results of IT/JT in 3-Joint datasets.}
  \resizebox{0.9\linewidth}{!}{
    \begin{tabular}{c|ccc}
    \hline
    \multicolumn{1}{c|}{Methods} & SHA   & SHB   & QNRF  \\
    \hline
    Individual Training (IT) & \multicolumn{1}{c|}{60.6}  & \multicolumn{1}{c|}{8.8}   & 97.7  \\
    \hline\hline
    Joint Training (JT)    & 60.2 ($\textcolor[rgb]{1,0,0}{\bf{\downarrow}}) $ & 9.3 ($\textcolor[rgb]{0,0,1}{\bf{\uparrow}}$)   & 92.8 ($\textcolor[rgb]{1,0,0}{\bf{\downarrow}}$)   \\
    \hline
    \end{tabular}%
    }\vspace{-1em}
  \label{tab:JT}%
\end{table}%

In this paper, we propose the \textbf{\emph{Domain-specific Knowledge Propagating Network}} (DKPNet) for multi-domain joint learning, which intends to refine the propagated knowledge according to the domain-specific distributions and highlight all domains without bias. Specifically, a novel \emph{Variational Attention} (VA) technique is introduced for facilitating the domain-specific attention learning. Based on VA, the output attention distribution can be easily controlled by the latent variable. And we apply the Gaussian Mixture distribution as the prior in the proposed VA for multi-domain learning. Furthermore, as an extension to VA, the \emph{Intrinsic Variational Attention} (InVA) is proposed for handling the potential problems of overlapped-domains and sub-domains. VA and InVA both insist to provide domain-specific guidance for knowledge propagating, but start from coarse and intrinsic perspectives, respectively. 
In summary, the contributions of this paper are listed as follows:

\vspace{-0.8em}
\begin{itemize}
  \item DKPNet is proposed to learn a general and robust density estimating model for crowd counting by multi-domain joint learning, which can successfully prevent the model from just learning several dominant domains, and can consistently improve the performances over all the datasets.
  \vspace{-0.5em}
  \item VA/InVA are introduced to provide the domain-specific guidance for refining the propagating knowledge with the help of latent variable. To our knowledge, it is the first work to use variational learning in attention for crowd counting.
  \vspace{-0.5em}
  \item Extensive experiments have been conducted on several popular datasets, including ShanghaiTech A/B\cite{zhang2016single}, UCF-QNRF\cite{idrees2018composition} and NWPU\cite{wang2020nwpu}, and achieve the state-of-the-art performances on the MAE evaluation.
      \vspace{-0.8em}
\end{itemize}
\section{Related Work}
\textbf{Crowd Counting}: We review the recent works by the techniques they applied. Such techniques include multi-scale~\cite{zhang2016single, sam2017switching, sindagi2017generating, zhang17sacnn, shen2018crowd}, multi-task~\cite{zhang2015cross, huang2018body, liu2018leveraging}, attention~\cite{liu2019adcrowdnet, sindagi2019ha, zhang2019relational, zhang2019attentional, jiang2020attention, miao2020shallow}, perspective map~\cite{shi2019revisiting, yan2019persp, yang2020reverse}, GNN~\cite{luo2020hybrid}, loss functions~\cite{ma2019bayesian}\cite{wang2020distribution}, classification~\cite{xhp2019SDCNet} detection~\cite{liu2018decidenet, sam2020locate}, NAS~\cite{hu2020count} and others~\cite{sindagi2017generating, shi2018crowd, liu2018crowd, ranjan2018iterative, liu2019context, wan2019adaptive, wang2019learning, sindagi2019multi, shi2019counting, liu2019point, bai2020adaptive}. However, none of these works pay attention to the multi-domain learning in crowd counting.

\textbf{Cross-domain Learning in Crowd Counting}: Cross-domain learning can be categorized into one/few shot learning~\cite{hossain2019one, reddy2020few}, domain adaption~\cite{li2019coda, wang2019learning, gao2019feature}, \etc. \cite{reddy2020few} presents a meta-learning inspired approach to solve the few-shot scene adaptive crowd counting problem, and \cite{hossain2019one} further introduces one-shot scene-specific crowd counting. For domain adaption, CODA~\cite{li2019coda} performs adversarial training with pyramid patches from both source- and target-domain, so as to tackle different object scales and density distributions.
Wang~\etal~\cite{wang2019learning} release a large synthetic dataset (GCC), and propose SE Cycle GAN to bridge the domain gap between the synthetic and real data. Gao~\etal~\cite{gao2019feature} propose Multi-level Feature aware Adaptation (MFA) and Structured Density map Alignment (SDA) to extract domain invariant features and produce density maps with a reasonable distribution on the real domain.

\textbf{Multi-domain Learning}: Multi-domain learning aims at improving the performance over multiple domains. And it has been exploited in many fields~\cite{misra2016cross, xiao2016learning, rebuffi2017learning, guo2018multi, liu2019compact, xiao2020multi,Li_2021_CVPR,chen2018virtual}.
Despite of heavy research on multi-domain learning in these fields, there are few corresponding research works in crowd counting. The most relevant work to our method is ~\cite{marsden2018people}. \cite{marsden2018people} presents domain-specific branches embedded behind a fixed ImageNet~\cite{deng2009imagenet} classification network. And, a domain classifier is proposed to decide which branch to process the input image. As a result, the final performance is not good and limited by the hard assignment of branches. Moreover, the computation cost of this work is in linearly correlation with the number of domains. However, different from this work, DKPNet is light in parameters and computation, and is more flexible and general in both training and testing phases, leading to much better performances. 

\textbf{Variational Learning}: VAE\cite{kingma2013auto} has been widely explored and used in generative model families and is good at controlling the output distribution via the latent variable. Based on VAE, conditional-VAE\cite{sohn2015learning} proposes a deep conditional model for structured output prediction using Gaussian latent variables; $\beta$-VAE\cite{higgins2016beta} is proposed to use balancing term $\beta$ to control the capacity and the independence prior; $\beta$-TCVAE\cite{chen2018isolating} further extends $\beta$-VAE by introducing a total correlation term. All these methods aim at using variational learning for generating visually-good images. However, in this paper, we integrate the variational learning into the attention mechanism, and propose the Variational Attention for learning domain-specific attentions.
\vspace{-0.5em}
\section{Proposed Method}
\vspace{-0.5em}
In this section, we will first give the motivation of our method in Sec.\ref{sec_background}, then introduce the \emph{Variational Attention} (VA) and the \emph{Intrinsic Variational Attention} (InVA) modules in Sec.\ref{sec_deca} and Sec.\ref{sec_inca}, \emph{resp}. Finally provide the whole pipeline of \textbf{\emph{Domain-specific Knowledge Propagating Network}} (DKPNet) in Sec.\ref{sec_dkpnet}.
\vspace{-0.5em}
\subsection{Motivation}\label{sec_background}
\vspace{-0.5em}

As experimented in Tab.\ref{tab:JT}, optimizing a deep model by directly employing all the data coming from SHA, SHB and QNRF datasets gives rise to the problem of biased domain learning behavior\cite{chen2019energy,chen2019mixed}, i.e. the deep model prefers to mainly focusing on the learning of the dominant domains instead of all the domains. This will lead to confusions in model prediction when giving data from non-dominant domains, since these domains are not well learned. Obviously, this phenomenon is unsatisfying and the learned model is not what we want.

Considering the above fact that not all of the useful knowledge from these datasets could be captured, this paper tries to use the \emph{attention mechanism}. \textbf{\emph{``Attention'' gives chances of capturing the desired information by re-weighting or refining the information/knowledge flow within deep models, so as to enhance the learning ability}}. However, the conventional attention modules like SENet\cite{hu2018squeeze,chen2019hybrid} are always ``\emph{self-attention}'' in fact. This will lead to the unconstrained and confused attention distribution outputs for different domains, and when these outputs are used again for re-weighting the original input data, the data distributions will be perturbed, resulting in difficulty of learning especially in multi-domain cases. 
Therefore, inspired by VAE\cite{kingma2013auto},in order to \textbf{control the attention outputs for domains with different distributions, we propose the \emph{Variational Attention} technique}.
\subsection{Variational Attention}\label{sec_deca}
Without loss of generality, suppose we have multiple datasets $\mathcal{X}^{*}$ and they have tight and different distributions with each other, each dataset has multiple instances $X^{*}_{i}\in{\mathcal{X^{*}}}, i\in{[1,\ldots,N^{*}]}$, $N^{*}$ refers to the image number of datasets $\mathcal{X}^{*}$, and each dataset has been given a coarse label $l^{*}$ where $l^{*}\in{[0,\ldots,C-1]}$ and $C$ is the number of datasets. After feeding each image into the deep model, we can obtain a $3$-D tensor $x$ at certain layer for learning attention proposals $y$.

As aforementioned, in order to control the attention distribution $p_{\theta}(y)$ where $\theta$ is the model parameter, we follow the VAE idea by introducing the latent variable $z$ for controlling the distribution $p_{\theta}(y)$. Specifically, for modeling different-domain attention distributions, we maximizing the log-likelihood of the conditional probability $(p_{\theta}(y|x,l))$ as follows:
\vspace{-1em}
\begin{align}
\log(p_{\theta}(y|x,l))&=\log(\int\frac{p_{\theta}(y,z|x,l)}{q_{\phi}(z|x,l)}q_{\phi}(z|x,l)dz)\nonumber\\
&\geq\mathbb{E}_{q_{\phi}(z|x,l)}\log(\frac{p_{\theta}(y|z,x,l)p_{\theta}(z|x,l)}{q_{\phi}(z|x,l)})\nonumber
\end{align}
\vspace{-0.5em}
\begin{align}\label{eq_vae}
  =\mathbb{E}_{q_{\phi}(z|x,l)}\log(p_{\theta}(y|z,x,l))-KL(q_{\phi}(z|x,l)||p_{\theta}(z|x,l))
\end{align}

This objective function is the evidence lower bound (ELBO) and  includes two terms. The first term tries to maximize the likelihood to improve the confidence of prediction, in other words, it tries to produce good attention proposals so as to benefit the density estimation (in this paper, it corresponds to density estimating loss, which will be described later, $\int_{q_{\phi}(z|x,l)}\|\hat{Y}(y)-Y\|^{2}_{2}dz$, where $Y$ is the ground-truth of density map and $\hat{Y}(y)$ is estimating result based on the attention output $y$). The second term refers to the KL divergence between the variational distribution $q_{\phi}(z|x,l)$ ( parameterized by $\phi$) and the prior distribution $p_{\theta}(z|x,l)$, as it is prior distribution we use $p(z|x,l)=p_{\theta}(z|x,l)$ later. Since the output attention distributions for different domains should be different with each other, we set the prior distribution of the latent variable $z\in\mathbb{R}^{d}$ to a commonly used Gaussian mixture distribution with $C$ Gaussian components, where $C$ is the number of domains and $d$ is the dimension of $z$:
\vspace{-0.5em}
\begin{align}\label{eq_gmm}
z\sim \sum_{c=0}^{C-1}\gamma_{c}\mathcal{N}(u_{c},\Sigma_{c}),~\forall{c},\gamma_{c}\geq0,\sum_{c=0}^{C-1}\gamma_{c}=1
\vspace{-0.5em}
\end{align}
For each domain, $u_{c}$ is the corresponding mean vector and $\Sigma_{c}$ is the $d$-dimensional covariance matrix. For convenience, we set $\gamma_{c}=\frac{1}{C}$ and $\Sigma_{c}$ is diagonal matrix throughout this paper. Then, the second term in Eq. \ref{eq_vae} can be expressed as:
\vspace{-0.5em}
\begin{align}\label{eq_kl}
&KL(q_{\phi}(z|x,l)||p(z|x,l))=\frac{1}{2}[\log(\frac{det(\Sigma_{c})}{det(\Sigma_{\phi})})-d\nonumber\\
&+tr(\Sigma_{c}^{-1}\Sigma_{\phi})+(u_{c}-u_{\phi})\Sigma_{c}^{-1}(u_{c}-u_{\phi})^{T}]
\vspace{-0.5em}
\end{align}
where $l=c$, the parameterized distribution $q_{\phi}(z|x,l)\sim\mathcal{N}(u_{\phi},\Sigma_{\phi})$, and $u_{\phi},\Sigma_{\phi}$ are the outputs of model $\phi$.

As the $\mathbb{E}_{q_{\phi}(z|x,l)}log(p_{\theta}(y|z,x,l))$ is computationally intractable and the total process should be differentiable, we use the reparameterization trick \cite{kingma2013auto} for computation as:
\vspace{-0.5em}
\begin{small}
\begin{align}\label{eq_reparam}
\mathbb{E}_{q_{\phi}(z|x,l)}\log(p_{\theta}(y|z,x,l))\simeq\frac{1}{N}\sum_{j=1}^{N}\log(p_{\theta}(y|z^{j},x,l))
\vspace{-1em}
\end{align}
\end{small}
where $z^{j}$ is sampled by $u_{\phi}+\Sigma_{\phi}\odot\varepsilon,\varepsilon\sim \mathcal{N}(0,I)$.
\begin{figure}[!t]
  \centering
  \includegraphics[width=0.9\linewidth]{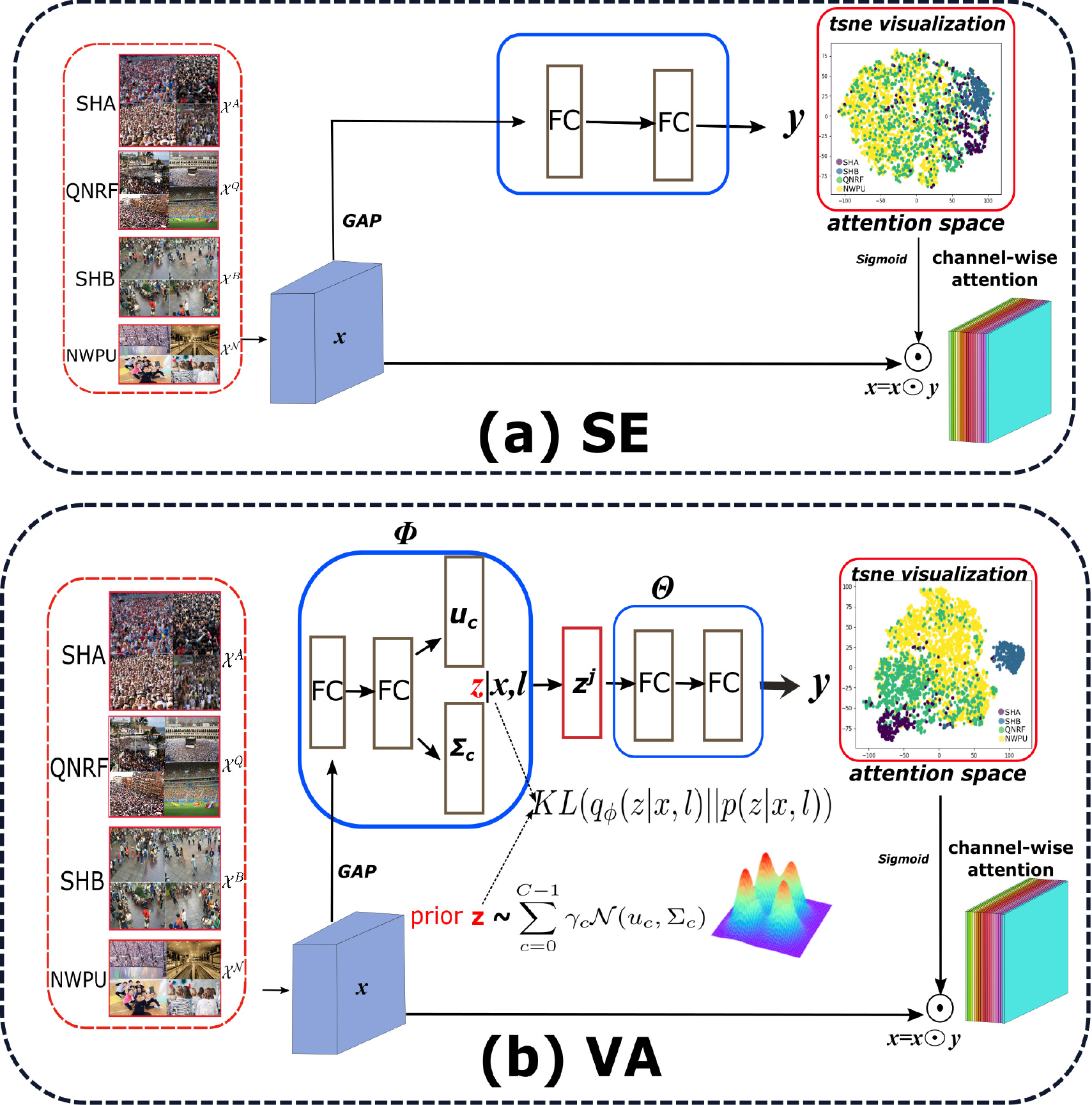}\\
  \caption{Comparisons between SE and VA. Different datasets are trained jointly. $\odot$ means channel-wise product. One can observe that SE attention outputs are confusing, while our VA can produce more separable attention distributions for different domains by introducing the Gaussian Mixture distributed latent variable $z$.}\label{fig_deca}
  \vspace{-1.2em}
\end{figure}

As mentioned before, each domain is supposed to have a Gaussian distribution, however, it is hard to factitiously set and fix the prior parameters $u_{c},\Sigma_{c}$. We propose to make them to be the learnable parameters for adaptively adjusting and add a distribution regularizer for getting non-trivial solutions. Considering the semantic relations between domains, we propose to regularize the similarity between the sampled $z^{j}$ coming from $c$-th domain and the prior learnable parameters $u_{c}$ to be the largest among all the similarities between $z^{j}$ and $u_{i},i\in[0,\ldots,C-1]$. This can be expressed to force:
\vspace{-1em}
\begin{align}\label{eq_sim}
{z^{j}}^{T}u_{c} &\geq \{{z^{j}}^{T}u_{0},\ldots,{z^{j}}^{T}u_{C-1}\}\Rightarrow \nonumber\\ \max({z^{j}}^{T}u_{0}-&{z^{j}}^{T}u_{c},\ldots,{z^{j}}^{T}u_{C-1}-{z^{j}}^{T}u_{c})\leq0
\vspace{-1em}
\end{align}
because the above $\max$ function is piecewise discontinuous function, here, we minimize its upper-bound function $\log$-$sum$-$\exp$ instead for optimization:
\vspace{-1em}
\begin{align}\label{eq_logsumexp} L_{reg}= \log(\sum_{i=0}^{C-1}e^{{z^{j}}^{T}u_{i}-{z^{j}}^{T}u_{c}})&\geq \max({z^{j}}^{T}u_{0}-{z^{j}}^{T}u_{c},\nonumber\\
\vspace{-1em}
\ldots,{z^{j}}^{T}&u_{C-1}-{z^{j}}^{T}u_{c})
\vspace{-0.5em}
\end{align}
Minimizing Eq.\ref{eq_logsumexp} can help regularizing different domains to have different distributions. And for $\Sigma_{c}$, we simply regularize it by minimizing $\log^{2}(det(\Sigma_{c}))$, and add it to $L_{reg}$. Finally, VA can be optimized by minimizing the following loss:
\vspace{-1em}
\begin{align}\label{eq_va}
L_{VA}&=\frac{-1}{N}\sum_{j=1}^{N}\log(p_{\theta}(y|z^{j},x,l))+ KL(q_{\phi}(z|x,l)||p(z|x,l))\nonumber\\
&+\log(\sum_{i=0}^{C-1}e^{{z^{j}}^{T}u_{i}-{z^{j}}^{T}u_{c}})+ \log^{2}(det(\Sigma_{c}))
\vspace{-1em}
\end{align}
and the attention output $y$ will be further used to re-weight the input tensor $x$ as $x=x\odot y$.

\textbf{Remark}: By introducing and modeling the latent variable, the output attention will be domain-related, such that the domain-specific knowledge can be well captured and learned in multi-domain cases. As shown in Fig. \ref{fig_deca}, different from SE attention, our VA can produce more separable attention distributions for different domains with the help of latent variable modeling. And the latent variable is assumed to be Gaussian Mixture distribution, such that for each domain, an independent Gaussian distribution can be applied and used to control the attention outputs.
\subsection{Intrinsic Variational Attention}\label{sec_inca}
Actually, the above VA simply assumes that each dataset belongs to a single domain. However, it does not hold on in many cases and there might be two frequent problems: (1) \emph{domain-overlaps} across different datasets and (2) \emph{sub-domains} within the same dataset. Therefore, labels $l$ will be too coarse to provide the fine-grained and exact guidance for domain-specific attention learning, still leaving behind some confusions in attention learning. To this end, in order to capture the intrinsic domain labels for accurate domain-specific attention learning, we extend our VA into \textbf{\emph{Intrinsic Variational Attention}} (InVA). It is mainly achieved by using \emph{Clustering}(CL) labels and \emph{Sub-Gaussian Components}(SGC) which intend to mitigate the problems of domain-overlaps and sub-domains, respectively.

Specifically, in order to tackle the domain-overlaps, it is required to reassign the more correct domain-labels to the training data instead of using the original dataset-label $l$. Thus, we first train a VA and then perform the Gaussian-Mixture clustering\footnote{We also tested other clustering methods, e.g. Kmeans, DBSCAN, \emph{etc}, the performances are similar.} over the attention proposals outputed by the VA module. The number of clusters is set to $\bar{C}$. 
After clustering, the new domain labels $\bar{l}\in{[0,\ldots,\bar{C}-1]}$, i.e. CL labels, are reassigned to the original training data.
\begin{figure*}[!t]
\centering
\vspace{-0.5em}
\includegraphics[width=0.9\linewidth]{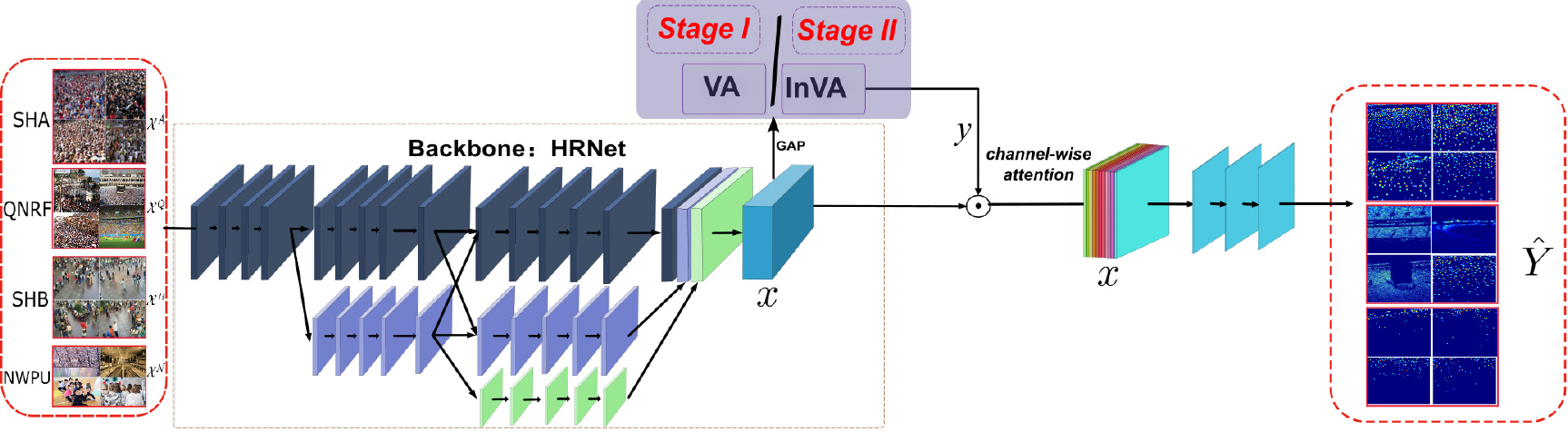}\\
\caption{The pipeline of the proposed DKPNet. It contains two-stage training: VA and InVA are used in Stage-I and Stage-II, respectively. Before Stage-II training, we will first get the CL labels from the attention outputs of VA by clustering, and then reassign the training images with these CL labels.}\label{fig_dkpnet}
\vspace{-1.5em}
\end{figure*}

Moreover, there also might be sub-domains within each clustered domain since the clustering is unsupervised and not capable of finding each potential sub-domains. In order to cope with the potential sub-domains, we propose to use SGC for the latent variable modeling. Specifically, we assume there are at most $k$ sub-domains in each clustered domain. Thus, the latent variable $z\in\mathbb{R}^{d}$ will turn to be a Sub-Gaussian Mixture distribution as follows:
\vspace{-0.5em}
\begin{align}\label{eq_sgmm}
z&\sim \sum_{c=0}^{\bar{C}-1}\frac{1}{\bar{C}}\mathcal{N}(u_{c},\Sigma_{c})\nonumber\\
where~u_{c}=\arg&\max_{u_{c,i}}\{\sigma(u_{c,1}^{T}u_{\phi}),\ldots,\sigma(u_{c,k}^{T}u_{\phi})\}
\vspace{-1em}
\end{align}
$\sigma$ means Dropout with drop-rate $0.2$, $u_{c,k}$ is center vector of the $k$-th sub-component in the $c$-th Gaussian. For simplicity, we use the same $\Sigma_{c}$ for the sub-gaussian components. Finally, using the CL labels $\bar{l}$ and substituting the latent variable $z$ in Eq. \ref{eq_sgmm} into Eq. \ref{eq_va} can obtain the loss function $L_{InVA}$ for training InVA.


\textbf{Remark}: CL labels focus on handling the overlapped-domains across different datasets. SGC allows the existence of sub-domains in each clustered domain and is capable of adaptively optimizing these sub-domains. Thus, equipping with both CL labels and SGC, InVA module can provide more accurate guidance for knowledge refining by attention according to the ``\emph{intrinsic}'' domains of data. As the basic module structure is similar with VA except for the CL labels and prior distribution of $z$, here for the limitation of paper length, we omit the figure for showing InVA.
\vspace{-0.5em}
\subsection{DKPNet}\label{sec_dkpnet}
Now, we will introduce the whole pipeline of our DKPNet as in Fig.\ref{fig_dkpnet}. Here, we take the truncated HRNet\cite{wang2020deep}(for parameter lightness, we only use parameters from \emph{stage1} to \emph{stage3}) as our backbone. During training, the mini-batched images are randomly sampled from all datasets and then are fed into the backbone together. A $1\times1$ convolution with $512$ channels is applied after, producing a $4$-D tensor $x$. Then tensor $x$ will be passed to VA/InVA for producing the domain-specific attention weights $y$, which will be applied on $x$ by channel-wise attention, resulting in a new tensor $x=x\odot y$. Notably, VA/InVA are used in stage-I and stage-II, respectively. And for each stage training, the backbone is initialized by the ImageNet-pretrained model. Finally, for predicting the density maps $\hat{Y}(y)$, three $1\times1$ convolution layers (with channels $64,32,1$, respectively) are adopted. After replacing $-\log(p_{\theta}(y|z^{j},x,l))$ by the density estimating loss $\|\hat{Y}(y)_{i}-Y_{i}\|^{2}_{2}$, the whole DKPNet can be optimized by the following objective function:
\vspace{-1em}
\begin{align}\label{eq_dkpnet}
  L =&\frac{1}{2B}\sum_{i=1}^{B}\|\hat{Y}(y)_{i}-Y_{i}\|^{2}_{2} + KL(q_{\phi}(z|x,l)||p(z|x,l))\nonumber\\
&+\log(\sum_{i=0}^{C-1}e^{{z^{j}}^{T}u_{i}-{z^{j}}^{T}u_{c}})+ \log^{2}(det(\Sigma_{c}))
  \vspace{-1.5em}
\end{align}
$B$ is the batch size and for VA/InVA, the latent variable $z$ are modeled by Eq. \ref{eq_gmm} and Eq. \ref{eq_sgmm}, \emph{resp}.

\textbf{Remark}: VA/InVA are performed in order, and progressively aim at refining the propagating information flows by attention, such that the data from different distributions can be unbiasedly treated and learned, without inducing confusions in predictions.
\vspace{-1em}
\section{Experiments}
\begin{table*}[!thbp]
    \caption{Results on SHA, SHB, QNRF and NWPU. ``Individual'' means the models are trained by only one individual dataset. ``3-Joint'' and ``4-Joint'' refer to using the joint dataset of (SHA,SHB,QNRF) and (SHA,SHB,QNRF,NWPU), respectively. For each joint dataset, only a single model is trained. Moreover, ``IT'' refers to training models just with the individual datasets. ``JT'' means merging all the datasets for training. For fair comparison, IT and JT are performed under the same training settings as our DKPNet. The best results are in \textcolor[rgb]{1,0,0}{red} color. NWPU(V) and NWPU(T) indicate Val and Test sets on NWPU, \emph{resp}.}
    \centering
  \resizebox{0.85\linewidth}{!}{
    \begin{tabular}{c|cc|cc|cc|cc|cc}
    \hline
    \multicolumn{11}{c}{\textbf{Training Dataset: Individual}} \\
    \hline
    \multirow{2}[2]{*}{Methods} & \multicolumn{2}{c|}{SHA\cite{zhang2016single}} & \multicolumn{2}{c|}{SHB\cite{zhang2016single}} & \multicolumn{2}{c|}{QNRF\cite{idrees2018composition}} & \multicolumn{2}{c|}{NWPU\cite{wang2020nwpu} (V)} & \multicolumn{2}{c}{NWPU\cite{wang2020nwpu} (T)}  \\
          & MAE   & MSE   & MAE   & MSE   & MAE   & MSE   & MAE   & MSE & MAE   & MSE\\
    \hline
	CSRNet~\cite{li2018csrnet} & 68.2 & 115.0  & 10.6 & 16.0 & - & -  & 104.8 & 433.4 & 121.3 & 387.8  \\
	CANet~\cite{liu2019context} & 62.3 & 100.0 & 7.8 & 12.2 & 107.0 & 183.0  & 93.5 & 489.9 & 106.3 & 386.5 \\
	SFCN~\cite{wang2019learning} & 64.8 & 107.5 & 7.6 & 13.0 & 102.0 & 171.0 & 95.4 & 608.3 & 105.4 & 424.1 \\
	DSSINet~\cite{liu2019crowd} & 60.6 & 96.0 & 6.9 & \textcolor[rgb]{1,0,0}{10.3} & 99.1 & 159.2  & - & - & - & -  \\
    Bayes~\cite{ma2019bayesian} & 62.8  & 101.8 & 7.7   & 12.7  & 88.7  & 154.8 &  93.6 & 470.3 & 105.4 & 454.2\\
    DM-Count~\cite{wang2020distribution} & 59.7  & 95.7  & 7.4   & 11.8  & 85.6  & 148.3 & 70.5  & \textcolor[rgb]{1,0,0}{357.6} & 88.4 & 388.6\\
    \hline\hline\hline
    \emph{IT}(baseline1) & 60.6  & 99.2 & 8.8  & 12.6 & 97.7 & 155.7 & 81.7  & 516.0 & 94.0 & 371.9\\
    \hline
    \multicolumn{11}{c}{\textbf{Training Dataset: 3-Joint}} \\
    \hline
    \emph{JT}(baseline2)    & 60.2  & 99.6  & 9.3     & 13.7 & 92.8  & 159.7 &    -   & - & - & -\\
    MB~\cite{marsden2018people}    & 59.4  & 101.2  & 8.3     & 13.2 & 91.9  & 159.6 &    -   & - & - & -\\
    \textbf{\emph{DKPNet (c=3,k=3)}} & \emph{\textbf{56.7}} & \emph{\textbf{97.1}}  & \emph{\textbf{6.9}}  & \emph{\textbf{12.0}} & \emph{\textbf{85.2}}  & \emph{\textbf{151.4}} &  -     & - & - & -\\
    \hline
    \multicolumn{11}{c}{\textbf{Training Dataset: 4-Joint}} \\
    \hline
    \emph{JT}(baseline2)    & 59.9  & 96.7  & 9.7     & 15.2  & 91.1  & 160.4 & 73.2  & 509.5 & 81.9 & 351.5\\
    MB~\cite{marsden2018people}    & 59.2  & 97.7  & 8.9     & 13.4 & 90.6  & 157.1 &    72.7   & 504.0 & 80.5 & 377.8\\
    \textbf{\emph{DKPNet (c=5,k=2)}} & \textcolor[rgb]{1,0,0}{\textbf{\emph{55.6}}} & \textcolor[rgb]{1,0,0}{\textbf{\emph{91.0}}}  & \textcolor[rgb]{1,0,0}{\textbf{\emph{6.6}}}  & \textbf{\emph{10.9}} & \textcolor[rgb]{1,0,0}{\textbf{\emph{81.4}}} & \textcolor[rgb]{1,0,0}{\textbf{\emph{147.2}}} & \textcolor[rgb]{1,0,0}{\textbf{\emph{61.8}}} & \textbf{\emph{438.7}} & \textcolor[rgb]{1,0,0}{\textbf{\emph{74.5}}} & \textcolor[rgb]{1,0,0}{\textbf{\emph{327.4}}}\\
    \hline
    \end{tabular}%
    }\vspace{-1.5em}
  \label{tab:sota}%
\end{table*}%
\textbf{Datasets}: We conduct experiments on ShanghaiTech A/B\cite{zhang2016single}, UCF-QNRF\cite{idrees2018composition} and NWPU\cite{wang2020nwpu}. SHA contains $482$ crowd images with crowd numbers varying from $33$ to $3139$, where $300$ images are used for training and the rest $182$ images are used for testing. SHB contains $716$ images with crowd numbers varying from $9$ to $578$, where $400$ images are employed for training and the rest $316$ images are for testing. QNRF \cite{idrees2018composition} contains $1,535$ images. These images are split into the training set with $1,201$ images and the testing set with $334$ images, respectively. This dataset has much more annotated heads and prefers highly-congested density. NWPU dataset\cite{wang2020nwpu} is a new public dataset which consists of $5,109$ images, including $3,109$ training images, $500$ val images and $1,500$ test images, where the test images can only be evaluated on the official website. 

\textbf{Mention \& Notations}: As in \emph{multi-domain} joint learning, these datasets are trained simultaneously by a single model, and tested individually, i.e. reporting the results on each datasets separately. ``DKPNet$(c,k)$'' denotes that we use $c$ clustered domains and at most $k$ sub-domains in the proposed DKPNet.

\textbf{Implementation details}: The proposed DKPNet is applied on the truncated HRNet-W40\cite{wang2020deep} architecture which is pretrained on ImageNet\cite{deng2009imagenet}. For model training, we adopt the Adam\cite{kingma2014adam} optimizer with default betas$=(0.9,0.999)$, set the start learning rate to be 0.00005 for both the pretrained backbone and the new added layers, and use a total of 450 epochs with the learning rate
decreased by a factor of 2.5 per 250 epochs. For data preprocessing, we adopt fixed Gaussian kernels with size $15$ to generate the ground-truth density maps. For images with the shortest side smaller than $416$,  we will resize the shortest side to $416$ by keeping the aspect ratio. Then during training, random-cropping with size $400\times400$, random horizontal flipping and color jittering are adopted. We set the batch size to 32 in all experiments and use two NVIDIA-V100 GPUs. DKPNet is implemented by Pytorch\cite{pytorch} framework.

\textbf{Evaluation Metrics}: We adopt MAE and MSE metrics for evaluations on crowd counting datasets, which is consistent with previous work\cite{li2018csrnet}.
%
\subsection{Comparison with State-of-the-Arts}
In order to highlight the significance of the proposed DKPNet, we compare it with some recent remarkable works over the popular challenging benchmarks, inlcuding ShanghaiTech A/B\cite{zhang2016single}, UCF-QNRF\cite{idrees2018composition} and NWPU\cite{wang2020nwpu}. IT and JT are conducted with the same training configurations as our DKPNet, and both of them are set as our baselines.

\textbf{More Data \& More Better ?} To answer this question, we provide some results as in Tab.\ref{tab:sota}. From this table, one can observe that merging\footnote{In JT, we have used the balanced-sampling strategy for different datasets.} more datasets for training a robust estimating model is infeasible. For example (1) in 3-joint case, comparing JT with IT, \emph{JT will biasedly sacrifice the performances on SHB}, i.e. raising the MAE from 8.8 to 9.3 (the same phenomenon can also be observed in the 4-joint datasets training case, i.e. JT raises the MAE on SHB from 8.8 to 9.7). (2) comparing JT results in 3-Joint case with those in 4-Joint case, one can observe that the performances are similar in both 3-Joint and 4-Joint cases, and even the performance on SHB in 4-Joint case is worse than that in 3-Joint case.
From the above observations, we can get conclusion that \textbf{\emph{using more datasets cannot easily result in a much better model}}.

\begin{figure}[!tbp]
\vspace{-1em}
  \centering
  \includegraphics[width=1\linewidth]{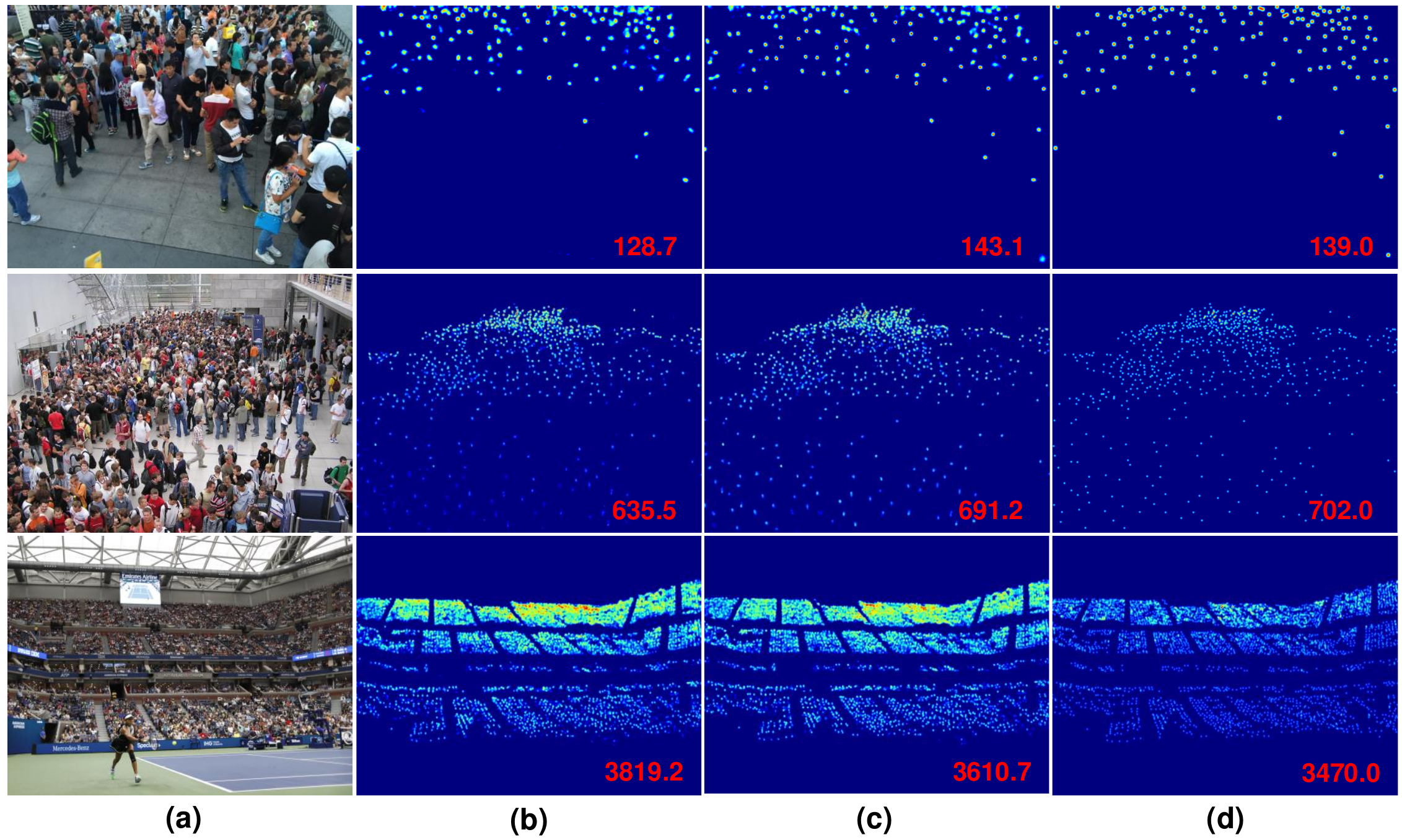}
  \vspace{-1em}
  \caption{Visualizations of test samples in the 4-Joint case. (a), (b), (c) and (d) are the input, density maps of JT, DKPNet(c=5,k=2) and Ground-Truth, \emph{resp}.}\label{fig_den}
  \vspace{-1.3em}
\end{figure}

\textbf{Effectiveness of DKPNet}: To this end, DKPNet is proposed and applied in both 3-joint and 4-joint cases. As in Tab.\ref{tab:sota}, one can observe that DKPNet can successfully use more data for learning a better model, i.e. it can consistently improve the performances over both IT and JT baselines. For example, (1) when using 3-Joint(or 4-Joint) datasets DKPNet can surpass the baseline IT by a large margin(e.g. obtaining 61.8 MAE on NWPU(V) with 24\% gains); (2) when using 4-Joint datasets, DKPNet can further improve the performances over DKPNet in 3-Joint case. These results verify the importance of our DKPNet for propagating domain-specific knowledge by using the latent-variable-constrained attention. Moreover, DKPNet(c=5,k=2) outperforms all the listed methods in MAE evaluation by a large margin, demonstrating the effectiveness of our DKPNet for multi-domain learning in crowd counting. Notably, DKPNet requires only \textbf{O}ne model for all the evaluations, while other methods have to require the corresponding trained model for each dataset. And the visualizations of density maps of JT and DKPNet are in Fig.\ref{fig_den}.

\textbf{Comparison with Multi-Branch Learning}: Moreover, comparing with the most related work MB\cite{marsden2018people} which uses a shared backbone followed by multiple branches for processing the different datasets(in which the parameter number and computational cost will increase with the number of datasets in a linear manner), DKPNet can surpass it by a large margin, as shown in Tab. \ref{tab:sota}, with negligible increases of parameters and computational costs. And MB uses the hard assignment of branches which is too ``arbitrary'' and cannot handle the problems of overlapped-domains and sub-domains, leading to a few performance improvements over the baseline JT. In contrast, DKPNet is much softer for handling the different data domains by using latent variable constrained attention, and is capable of handling these above problems.
\subsection{Component Analysis}
\begin{figure}[!hbp]
\vspace{-1em}
  \centering
  \includegraphics[width=1\linewidth]{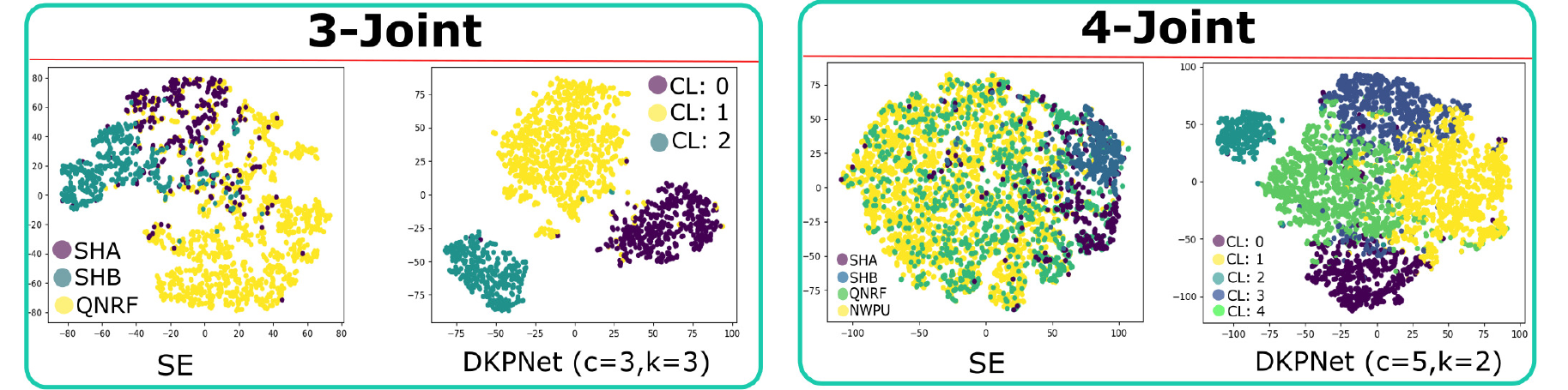}
  \centering
  \vspace{-1em}
  \caption{Tsne\cite{maaten2008visualizing} visualizations of the attention outputs of SE\cite{hu2018squeeze} and our DKPNet. Different colors refer to different domains based on the GT labels and CL labels.}\label{fig_se}
  \vspace{-0.5em}
\end{figure}
\begin{table}[!hbp]
  \centering\centering
  \vspace{-0.5em}
  \caption{MAE comparisons between SE\cite{hu2018squeeze} and DKPNet. For 3-Joint and 4-Joint cases, we use DKPNet(c=3,k=3) and DKPNet(c=5,k=2), respectively.}
  \resizebox{1\linewidth}{!}{
    \begin{tabular}{c|ccc|cccc}
    \hline
    \multicolumn{1}{c|}{Methods} & SHA   & SHB   & QNRF  & SHA   & SHB   & QNRF  & NWPU(V) \\
    \hline\hline
    IT & 60.6  & 8.8   & 97.7  & 60.6  & 8.8   & 97.7  & 81.7 \\
    \hline
          & \multicolumn{3}{c|}{\textbf{3-Joint}} & \multicolumn{4}{c}{\textbf{4-Joint}} \\
    \hline
    SE\cite{hu2018squeeze}    & 58.1  & 9.4   & 88.1  & 58.0    & 9.6     & 97.8  & 66.4 \\
    DKPNet & \textbf{56.7}  & \textbf{6.9}   & \textbf{85.2}  & \textbf{55.6}  & \textbf{6.6}   & \textbf{81.4}  & \textbf{61.8} \\
    \hline
    \end{tabular}%
    }\vspace{-1em}
  \label{tab:se}%
\end{table}%
\textbf{Why the domain-specific attention works?} In order to answer this question and to demonstrate the effectiveness of our enhanced attention method in DKPNet, we take the SE\cite{hu2018squeeze} attention for comparison. For fair comparison, we train the SE-based model via adopting the same training configurations and backbone as our DKPNet, only replacing the VA/InVA modules with the SE attention module. We first provide the comparisons of attention spaces via tsne visualization as in Fig.\ref{fig_se}. Specifically, as in the 3-Joint case, when training with the SE module, the attention distributions for different datasets are very close to each other and even confuse with each other, e.g. the attention weights for SHB are very close to those for SHA and QNRF, even overlaps with them. However, this confusing attention output does not bring the performance improvements on the SHB dataset (i.e. as shown in Tab.\ref{tab:se}, MAE result on SHB will be weakened from 8.8 to 9.4). It is because that the deep model will produce the accurate predictions if and only if it can exactly capture the true data distributions and then specifically learn knowledge from them. Therefore, when employing the SE attention which has no explicit domain guidance, the images in SHB (which have clear distribution differences with images in (SHA, QNRF)) will be wrongly assigned with the (SHA, QNRF)-like attention weights, resulting in further confusion in model prediction. Moreover, the same phenomenon can also be observed in the 4-Joint case, i.e. SE produces much more confused attention distributions, meanwhile it results in the performance drops on both SHB and QNRF, i.e. drop from 8.8 to 9.6 and from 97.7 to 97.8, \emph{resp}.

However, in contrast, as shown in Fig.\ref{fig_se} and Tab.\ref{tab:se}, when training with our domain-specific attention modules, the output attention spaces are much more separable than those outputted by SE. As a result, the propagating information can be handled specifically and exactly for producing correct predictions. Finally, the performances on all datasets can be consistently improved without biases, outperforming SE by a large margin. These phenomenons demonstrate the necessity and importance of explicitly learning the domain-specific attention for multi-domain joint training.

\begin{figure}[!tbp]
  \centering
  \vspace{-1.5em}
  \includegraphics[width=0.9\linewidth]{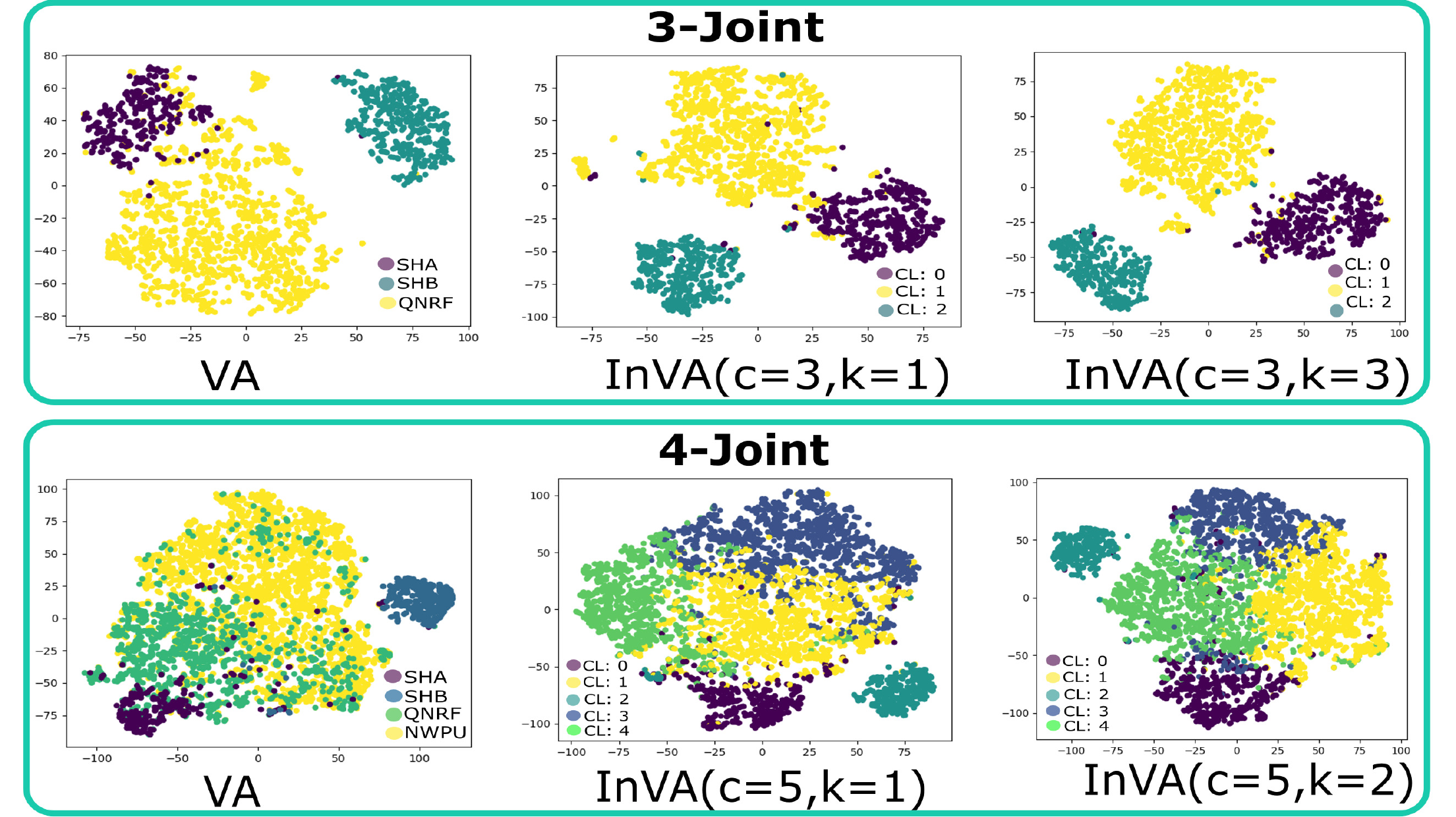}
  \centering
  \caption{Tsne visualizations of the attention outputs.}\label{fig_tsne_deca_ince}
  \vspace{-1em}
\end{figure}
\begin{table}[!tbp]
  \centering
  \caption{The MAE results comparisons of VA/InVA.}
  \resizebox{1\linewidth}{!}{
    \begin{tabular}{c|ccc|c|cccc}
    \hline
    \multicolumn{4}{c|}{\textbf{3-Joint}}   & \multicolumn{5}{c}{\textbf{4-Joint}} \\
    \hline
    Methods & SHA   & SHB   & QNRF  & Methods & SHA   & SHB   & QNRF  & NWPU(V) \\
    \hline
    IT    & 60.6  & 8.8   & 97.7  & IT    & 60.6  & 8.8   & 97.7  & 81.7 \\
    JT    & 60.2  & 9.3   & 92.8  & JT    & 59.9  & 9.7   & 91.1  & 73.2 \\
    VA    & 57.5  & 7.6   & 87.9  & VA    & 57.6  & 7.2   & 87.6  & 66.2 \\
    InVA(c=3) & 57.3  & 7.5   & 86.7  & InVA(c=5) & 56.5  & 7     & 84    & 63.9 \\
    InVA(c=3,k=3) & \textbf{56.7}  & \textbf{6.9}   & \textbf{85.2}  & InVA(c=5,k=2) & \textbf{55.6}  & \textbf{6.6}   & \textbf{81.4}  & \textbf{61.8} \\
    \hline
    \end{tabular}%
    }\vspace{-1.5em}
  \label{tab:va_inva}%
\end{table}%
\textbf{Effect of the VA and InVA}: As in Tab.\ref{tab:va_inva} and Fig.\ref{fig_tsne_deca_ince}, we conduct the quantitative and qualitative comparisons on DKPNet. For convenience, we will take the 3-Joint case for major description. Specifically, the VA tries to learn a coarsely domain-specific attention output via the latent variable $z$. This leads to the consistent performance improvements over both the baselines IT and JT (e.g. 57.5 vs. 60.2/60.6, 7.6 vs. 9.3/8.8 and 87.9 vs. 92.8/97.7 on SHA, SHB and QNRF respectively.). The learned attention distributions for different datasets are relatively separable except for some special cases (some overlapped distributions). Moreover, considering that the dataset-labels are not the exact definitions of the intrinsic data domains, we propose the InVA module to further explore the intrinsic domains by data clustering and fine-grained Gaussian Mixture distribution modeling. For example, when using InVA(c=3), one can observe that the aforementioned overlapped distributions can be properly handled to some extent, obtaining further consistent performance improvements over VA (e.g. improving from 57.5 to 57.3, from 7.6 to 7.5 and from 87.9 to 86.7 on SHA, SHB and QNRF, \emph{resp}.). Furthermore, InVA(c=3,k=3) is proposed to cope with the potential sub-domains within each clustered domain. One can observe that after giving chances of learning the potential sub-domains, the attention distributions of each clustered domains are more compact than before and the quantitative performances are further improved(e.g. improving the MAE from 57.3 to 56.7, 7.5 to 6.9 and 86.7 to 85.2 on SHA, SHB and QNRF, \emph{resp}). And the similar performance improvements can also be observed in the 4-Joint case.

In summary, DKPNet concentrates on progressively learning the domain-specific guided information flow by the two-staged training framework, where VA and InVA are performed in order.

\begin{table}[!tbp]
  \centering
  \caption{MAE results on the value of ($c,k$) in the 3-Joint case. For convenience, we will omit writing $k$ when $k=1$.}
  \resizebox{1\linewidth}{!}{
    \begin{tabular}{c|ccc|c|ccc}\hline
    \multicolumn{1}{c|}{Methods} & SHA   & SHB   & QNRF  & \multicolumn{1}{c|}{Methods} & SHA   & SHB   & QNRF \\
    \hline
    InVA(c=2) & \textbf{57.1}  & 7.9   & 87.5  & InVA(c=3,k=2) & 57.1  & 7.3   & 85.9 \\
    InVA(c=3) & 57.3  & \textbf{7.5}   & \textbf{86.7}  & InVA(c=3,k=3) & \textbf{56.7}  & \textbf{6.9}   & \textbf{85.2} \\
    InVA(c=4) & 59.6  & 7.7   & 88.2  & InVA(c=3,k=4) & 57.7  & 7.4   & 86.2 \\
    \hline
    \end{tabular}%
    }\vspace{-1.5em}
  \label{tab:ck}%
\end{table}%
\textbf{Ablation Study on the value of ($c,k$)}: As mentioned before, in InVA, we will first obatin the clustered domains that are separable to some extent in the macro-perspective, and then handle the potential sub-domains. The experimental results are shown in Tab.\ref{tab:ck}, one can observe that, for the 3-Joint case, $c=3$ works the best. It is reasonable since from a global view of the attention space output by VA (see Fig.\ref{fig_tsne_deca_ince}), $c=3$ can well separate the different distributions without much confusion. And for each clustered domain, $k=2,3,4$ works better than $k=1$ since the potential sub-domains can be explicitly learned, and we experimentally find $k=3$ is the best. Moreover, for the 4-Joint case, we experimentally find ($c=5,k=2$) works the best.

\begin{table}[!bp]
  \vspace{-1.5em}
\begin{minipage}{.49\linewidth}
  \centering
  \caption{Cosine similarities between sub-domain centers for the 3-Joint case. Sim$(q,p)$ means the cosine similarity between sub-centers $u_{c,q}$ and $u_{c,p}$.}
  \resizebox{1\linewidth}{!}{
    \begin{tabular}{c|ccc}
    \hline
          & CL-0  & CL-1  & CL-2 \\
          \hline
    Sim$(0,1)$   & 0.61  & 0.42  & 0.90 \\
    Sim$(0,2)$   & 0.75  & 0.72  & 0.78 \\
    Sim$(1,2)$      & 0.67  & 0.59  & 0.80 \\
    \hline
    avg-sim          &  0.68  & 0.58  &  0.83 \\
          \hline
    \end{tabular}%
    }
  \label{tab:c3k3}%
\end{minipage}
\begin{minipage}{.49\linewidth}
  \centering
  \vspace{-0.8em}
  \vspace{0.4em}
  \caption{The number of images in different sub-domains for the 3-Joint case. ``Sub-$p$'' is the $p$-$th$ sub-domain.}
  \resizebox{1\linewidth}{!}{
          \begin{tabular}{c|ccc}
          \hline
          & CL-0  & CL-1  & CL-2 \\
          \hline
    Sub-0 & 180   & 423   & 56 \\
    Sub-1 & 218   & 86   & 257 \\
    Sub-2 & 92   & 510   & 79 \\
    \hline
    \end{tabular}%
    }
  \label{tab:c3k3_branch}%
  \end{minipage}
  \vspace{-1.3em}
\end{table}%
\textbf{Sub-domain analysis}: In order to explicitly show the learning results of the sub-domains, we compute the cosine similarities between the center vectors of sub-domains (parameterized by $u_{c,k}$) and also calculate out the number of images in each sub-domains as shown from Tab.\ref{tab:c3k3}-Tab.\ref{tab:c3k3_branch}. For example, from Tab.\ref{tab:c3k3}, one can observe that the sub-domains in each clustered domain are specifically learned and different with each other since Sim$(q,p)$ shows there exists angle between sub-domain centers. Moreover, from Tab.\ref{tab:c3k3_branch}, it can be observed that all the sub-domains have its corresponding images, implying that the sub-domains indeed exist and are successfully learned. 

\textbf{Ablation Study on regularization term}:
As shown in Tab. \ref{tab:regula}, comparing with DKPNet, when training DKPNet without regularizing the learning of the prior distribution of $z$, the performances will be weakened a little. This shows the importance of term $L_{reg}$ for improving the learning of domain-specific attentions by regularizing the differences across domain distributions.
\begin{table}[!htbp]
  \centering
  \vspace{-0.5em}
  \caption{Ablation study on regularization term.}
  \resizebox{0.9\linewidth}{!}{
    \begin{tabular}{c|ccccc}
    \hline
    Methods   & SHA   & SHB   & QNRF  & NWPU(V)  \\
    \hline
         JT(baseline)& 59.9   & 9.7  & 91.1 & 73.2   \\
    \hline\hline
         DKPNet w/o $L_{reg}$& 57.3  & 7.5   & 86.9  & 64.2  \\
        DKPNet& \textbf{55.6}  & \textbf{6.6}   & \textbf{81.4}  & \textbf{61.8}   \\
    \hline
    \end{tabular}%
    }\vspace{-1em}
  \label{tab:regula}%
\end{table}%


\textbf{Model size}: From Tab.\ref{tab:pn}, one can observe that the total parameter number of DKPNet is fewer than \cite{ma2019bayesian,wang2020distribution}, but DKPNet can surpass them by a large margin, showing that DKPNet is indeed ``light and sweet''. Moreover, DKPNet only needs one single model for all data evaluations, while \cite{ma2019bayesian,wang2020distribution} have to train many corresponding models for evaluations which are heavy and complicated.
\begin{table}[!htbp]
  \centering
  \vspace{-0.5em}
  \caption{Model size comparisons. PN means Parameter Number(million). DKPNet is trained by the 4-Joint datasets.}
  \resizebox{1\linewidth}{!}{
    \begin{tabular}{c|c|ccccc}
    \hline
    Methods  & PN & SHA   & SHB   & QNRF  & NWPU(V) & NWPU(T) \\
    \hline
    Bayes\cite{ma2019bayesian}  & 20    & 62.8  & 7.7   & 88.7  & 93.6  & 105.4 \\
    DM-count\cite{wang2020distribution}  & 20    & 59.7  & 7.4   & 85.6  & 70.5  & 88.4 \\
    DKPNet  & \textbf{14}    & \textbf{55.6}  & \textbf{6.6}   & \textbf{81.4}  & \textbf{61.8}  & \textbf{74.5} \\
    \hline
    \end{tabular}%
    }\vspace{-1.5em}
  \label{tab:pn}%
\end{table}%
\section{Conclusion}
In this paper, we propose DKPNet for learning the robust and general density estimating model for crowd counting by multi-domain joint learning. Specifically, DKPNet is a two-stage training framework, where the VA module is used in Stage-I for coarsely guiding the domain-specific attention learning, and the InVA module is employed in Stage-II for exploring the intrinsic domains by handling both the problems of overlapped-domains and sub-domains, so as to provide more accurate guidance for domain-specific attention learning. Finally, extensive experiments have been conducted on four popular benchmarks to validate the necessity and effectiveness of our method.

\small{\textbf{Acknowledgments}: We hereby give specifical thanks to Alibaba Group for their contribution to this paper}
{\small
\bibliographystyle{ieee_fullname}
\bibliography{egbib}
}

\end{document}